%% file: sample-sigconf.tex
\documentclass[sigconf]{acmart}

\AtBeginDocument{%
  \providecommand\BibTeX{{%
    \normalfont B\kern-0.5em{\scshape i\kern-0.25em b}\kern-0.8em\TeX}}}

\usepackage{xspace}
\usepackage{graphicx}
\usepackage{algorithmic}

\newcommand{\ignore}[1]{}
\usepackage{adjustbox}

\usepackage{booktabs} 

\setcopyright{rightsretained}

\hypersetup{draft}
\begin{document}

\title{Selecting Miners within Blockchain-based Systems Using Evolutionary Algorithms for Energy Optimisation}



\author{Akram Alofi}
\affiliation{
\institution{School of Computer Science, University of Birmingham}
\city{Birmingham}
\country{United Kingdom}}
\affiliation{
\institution{Computer Science Department, Jamoum University College, Umm Al-Qura University}
\city{Jamoum}
\country{Saudi Arabia}}
\email{ama848@cs.bham.ac.uk}

\author{Mahmoud A. Bokhari}
\affiliation{
\institution{Computer Science Department, Taibah University}
\city{Medina}
\country{Saudi Arabia}}
\affiliation{
\institution{Optimisation and Logistics, School of Computer Science, The University of Adelaide}
\city{Adelaide}
\country{Australia}}
\email{mabokhari@taibahu.edu.sa}

\author{Robert Hendley}
\affiliation{
\institution{School of Computer Science, University of Birmingham}
\city{Birmingham}
\country{United Kingdom}}
\email{r.j.hendley@cs.bham.ac.uk}

\author{Rami Bahsoon}
\affiliation{
\institution{School of Computer Science, University of Birmingham}
\city{Birmingham}
\country{United Kingdom}}
\email{r.bahsoon@cs.bham.ac.uk}

\renewcommand{\shortauthors}{A. Alofi et al.}

\renewcommand{\shorttitle}{Selecting Miners within BC-based Systems Using EAs for Energy Optimisation}

\begin{abstract}
In this paper, we represent the problem of selecting miners within a blockchain-based system as a subset selection problem. We formulate the problem of minimising blockchain energy consumption as an optimisation problem with two conflicting objectives: energy consumption and trust. The proposed model is compared across different algorithms to demonstrate its performance.
\end{abstract}

%
%



\copyrightyear{2021}
\acmYear{2021}
\setcopyright{rightsretained}
\acmConference[GECCO '21 Companion]{2021 Genetic and Evolutionary Computation Conference Companion}{July 10--14, 2021}{Lille, France}
\acmBooktitle{2021 Genetic and Evolutionary Computation Conference Companion (GECCO '21 Companion), July 10--14, 2021, Lille, France}\acmDOI{10.1145/3449726.3459558}
\acmISBN{978-1-4503-8351-6/21/07}

\maketitle

\input{samplebody-conf}

\bibliographystyle{ACM-Reference-Format}
\bibliography{sample-bibliography} 
\end{document}

%% file: samplebody-conf.tex
\section{Introduction}
Blockchain technology is a novel form of replicated database (`distributed ledger') that operates autonomously without a centralised control. It has several key characteristics, such as auditability, anonymity, persistence and decentralisation. Although this technology holds much promise for the future, it also has some challenges. A major issue is the energy consumption of the Proof of Work (PoW) consensus algorithm. PoW consumes a large amount of energy and will have significant environmental consequences if it is widely employed. In November 2019, a transaction of the most common blockchain-based system (Bitcoin~\cite{nakamoto2008bitcoin}) required on average 431 kWh of electricity which is enough energy to power 21 US homes for 24 hours~\cite{8972381}. 

Since PoW based blockchain-based systems employ considerable computing resources (and thus have considerable energy consumption), many researchers and organisations have proposed alternative consensus algorithms, in an effort to reduce its energy consumption. However, minimising energy consumption has not been formulated as an optimisation problem. In this work, we formulate the problem of selecting miners for mining blocks in a blockchain-based system as a subset selection problem. 

The problem is represented as selecting a set of miners within a blockchain network, where each miner demands a level of energy and has a level of reputation. Given the conflicting goals of minimising energy and maximising trust, we use evolutionary algorithms (EAs) to select miners that consume less energy and have high reputation values. The fitness function considers energy versus reputation. 

\section{Optimisation Problem Formulation} \label{sec:opt}
Blockchain-based systems are like many other real-world applications where there are trade-offs. In blockchain-based systems, the many objectives show the advantages of compromise, which can be considered one sort of optimisation problem. These objectives include, among others, energy consumption, trust, decentralisation, scalability, performance and security. We have summarised some conflicting blockchain objectives that can be utilised to optimise blockchain-based systems (See Table \ref{tab:ExamplesObjs}).

\begin{table}
  \caption{Examples of Blockchain Objectives for Optimisation Models}\vspace{-2 mm}
  \label{tab:ExamplesObjs}
  \resizebox{\columnwidth}{!}{%
  \begin{tabular}{lccc}
    \toprule
        & \textbf{Environmental} 
        & \textbf{Security} 
        & \textbf{Performance}\\
    \midrule
     Mining Device & \cite{TRUBY2018399} & \cite{REN2020161} & \cite{REN2020161} \\
     Number of Nodes & \cite{8720159} & \cite{8720159} & \cite{8720159} \\
     Consensus Algorithm & \cite{TRUBY2018399} & \cite{app9224731} & \cite{app9224731} \\
     Blockchain Type & \cite{casino2019systematic} & \cite{app9224731} & \cite{BAMAKAN2020113385} \\ 
     \bottomrule
    \end{tabular}
    }
    \vspace{-5 mm}
\end{table}

There have been several interesting and successful implementations of multi-objective optimisation evolutionary algorithms (MOEAs) for finding optimum and near-optimum solutions for different real-world application problems such as optimising the energy use of software~\cite{Bokhari2018mobiquitous, bruce:energy}. However, MOEAs have not been used to optimise for energy use and trust in blockchain-based systems. 

\subsection {Optimisation Model}\label{sec:fitnessModles}
In this work, we utilise MOEAs to improve the energy consumption of blockchain-based systems. We formulate the problem of minimising the energy consumption of a blockchain-based system by selecting a subset of miners, which decreases the energy use and maximises the trust level of the system; miners selection seeks high reputation values.

\subsubsection{Energy Consumption Objective} 
The focus of this work is on saving energy expended by miners during computing procedures to boost the sustainability of blockchain-based systems. The energy consumed during mining procedures accounts for a large proportion of the energy consumed by blockchain-based systems. Reducing this energy consumption level is the optimisation objective: the smaller the energy value, the fitter the solution. We calculate the total energy of each miner based on the energy consumption of the devices used during 24 hours.

\subsubsection{Reputation Objective}
Our model, which was inspired by PoS and PoW, features a reduced number of miners, which means we need to increase the blockchain-based systems' trust levels to aid the PoW consensus algorithm. The trustworthiness of miners within a blockchain network is evaluated after each published block. Thereafter, the reputation values are calculated using two features: each miner's stake and the number of blocks added by each miner are collected and used to calculate the reputation values.

\section{Evaluation} \label{sec:Exp}
To evaluate our proposal, we use a blockchain simulator (Bitcoin-Simulator \cite{10.1145/2976749.2978341}) to collect miners' data (the number of blocks produce by each miner and the rewards and fees miners earn). The simulator was set to simulate the behaviour of 160 miners for mining 4073 blocks. To run our experiments, we use the MOEA framework \footnote{MOEA Framework version 2.13 available at \url{http://moeaframework.org}, accessed 10 December 2020.}. Since we address the optimisation of the energy consumption for selecting miners as a new optimisation problem, we integrated our proposed fitness functions into two EAs (SPEA2 and NSGAII). We also use the Random Search (RS) algorithm as a baseline for our comparisons. Each algorithm was run 100 times with an evaluation budget of 40,000 per run. To approximate the Pareto front, we combine each algorithm's non-dominated set of solutions. 

\begin{figure}[t]
\includegraphics[width=\linewidth, height=3.5cm]{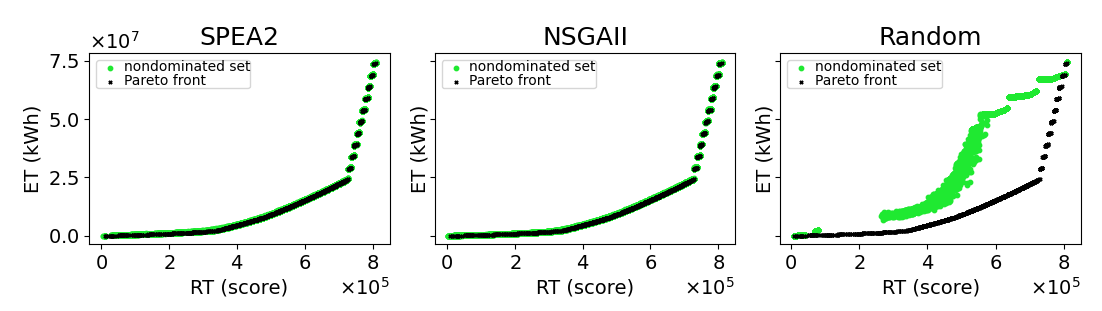}
\vspace{-7 mm}
\caption{Energy vs. Reputation}\label{fig:energyTrust}
\vspace{-5 mm}
\end{figure}

Figure \ref{fig:energyTrust} shows that SPEA2 and NSGA-II consistently find non-dominated solutions from the Pareto front. For minimising the energy use and maximising the reputation, we can clearly see that the RS is the worst performing among the algorithms. 

The results of the statistical test (two-tailed Wilcoxon rank-sum test ~\cite{wilcoxon1992individual}) and the effect size (Vargha and Delaney effect size~\cite{varghaDelanay}) for the performance of each algorithm show that the hypervolume of the Random Search algorithm is significantly lower than the other algorithm hypervolume. Of all the algorithms, NSGAII performed the best. The algorithm generated the most diverse, non-dominated set, which covers most of the search space.

For future work, we plan to conduct further experiments and investigations such as using different MOEAs and comparing their performance, conducting sensitivity analysis on the algorithms configurations. We will integrate our approach with the common consensus algorithm PoW and their variants to report on the effectiveness of the integration on its energy efficiency with several dimensions such as mining and energy source (e.g. renewable).

